# Knowledge-Guided Vision–Language Inference for Image-Based Urban Flood Depth Estimation

**Zhangding Liu[1]; Neda Mohammadi[2]; and John E. Taylor[3]**

[1]PhD Student, School of Computational Science and Engineering, Georgia Institute of Technology, 790 Atlantic Dr NW, Atlanta, GA 30332, United States; e-mail: zliu952@gatech.edu
[2]School of Civil and Environmental Engineering, Georgia Institute of Technology, 790 Atlantic Dr NW, Atlanta, GA 30332, United States; e-mail: nedam@gatech.edu; The University of Sydney, Sydney, NSW 2006, Australia; e-mail: neda.mohammadi@sydney.edu.au
[3]Professor, School of Civil and Environmental Engineering, Georgia Institute of Technology, 790 Atlantic Dr NW, Atlanta, GA 30332, United States; e-mail: jet@gatech.edu (corresponding author)

## ABSTRACT

Timely floodwater depth estimates support road accessibility assessment and emergency response during urban flooding. Supervised vision methods often require extensive labeled datasets, while recent foundation vision–language models (VLMs) offer flexible visual reasoning but can inconsistently yield large errors in metric depth estimation. This paper proposes FloodVision, a knowledge-guided framework for estimating flood depth from a single RGB image. FloodVision integrates a general-purpose VLM with FloodKG, a domain knowledge base encoding canonical object dimensions and component landmarks (e.g., wheel arch, curb top) to encourage reasoning at the component level rather than treating objects as wholes. This injects explicit geometric grounding without task-specific training. Evaluated on 654 crowdsourced MyCoast New York flood images with resident-reported depths as proxy labels, FloodVision reduces the mean absolute error from 15.62 cm to 8.75 cm and the median error from 14.35 cm to 7.75 cm, with lower error than the VLM-only baseline in 69.3% of cases. The paper also discusses current limitations and future integration into urban digital twin systems.

## INTRODUCTION

Urban floods are among the most frequent and destructive natural disasters, causing billions of dollars in infrastructure damage, disrupting transportation networks, and endangering public safety (Yadav et al. 2020). Flash floods can block roads, damage critical services, and disrupt utilities—for example, heavy rainfall in London triggered cascading failures in transport and power networks, exposing urban lifeline vulnerabilities (Yadav et al. 2020). Timely and accurate estimation of flood depth is therefore essential for real-time road accessibility mapping, flood damage assessment, and effective emergency response (Liu et al. 2025; Pan et al. 2023).

Existing urban flood depth estimation methods face limitations for urban applications. Traditional field surveys provide high accuracy but are time-consuming and unsuitable for rapid response. In-situ water level sensors provide real-time data but are costly to calibrate, sparsely distributed, and prone to failure during extreme events or in ungauged areas, which limits their ability to capture urban flood variability. Hydrodynamic models extend spatial coverage but require intensive calibration and high-resolution terrain data, making them computationally expensive for time-sensitive deployment (Athira et al. 2023). Recent advances in computer vision have enabled automated flood detection, but depth estimation methods often rely on detecting partially submerged, predefined reference objects like vehicles or street signs (Chaudhary et al. 2019; Li et al. 2023). These approaches face fundamental limitations: their dependence on large, annotated datasets restricts scalability, while their reliance on a fixed set of object detectors prevents them from generalizing to diverse urban scenes where such objects may be occluded or absent.

Foundation VLMs have been leveraged in disaster response for high-level semantic tasks like scene classification and damage assessment (Kane and Khose 2022). However, their application to quantitative flood depth estimation remains limited because numeric depth requires geometric grounding and consistent use of reference cues, whereas VLMs tend to provide qualitatively plausible but metrically inconsistent estimates when prompted directly.

To address this limitation, we propose FloodVision, a knowledge-guided vision–language framework for urban flood depth estimation from single RGB images. FloodVision integrates a general-purpose VLM with FloodKG, a lightweight knowledge base encoding canonical object dimensions and flood-relevant components. By mapping detected objects to canonical types and guiding the VLM to reason at the component level, FloodVision introduces explicit geometric grounding into VLM-based inference without task-specific training. On 654 crowdsourced MyCoast New York flood images, FloodVision reduces the mean absolute error (MAE) from 15.62 cm to 8.75 cm and the median error from 14.35 cm to 7.75 cm compared to a VLM-only baseline, demonstrating improved robustness for real-world flood monitoring.

## METHODOLOGY

This section presents FloodVision, a knowledge-guided vision–language framework for estimating urban floodwater depth from images (Figure 1). Instead of relying on large annotated datasets, FloodVision couples a general-purpose VLM with a compact domain knowledge base that provides geometric priors for common urban objects. The framework follows a three-stage pipeline. In Stage 1, the VLM performs visual semantic detection, identifying reference objects in the image and producing initial object-level depth estimates. In Stage 2, FloodKG maps detected objects to canonical types and retrieves components located near the waterline estimated in Stage 1. In Stage 3, the VLM re-analyzes the image under this knowledge guidance, focusing on selected components to generate refined flood depth estimates and structured outputs.

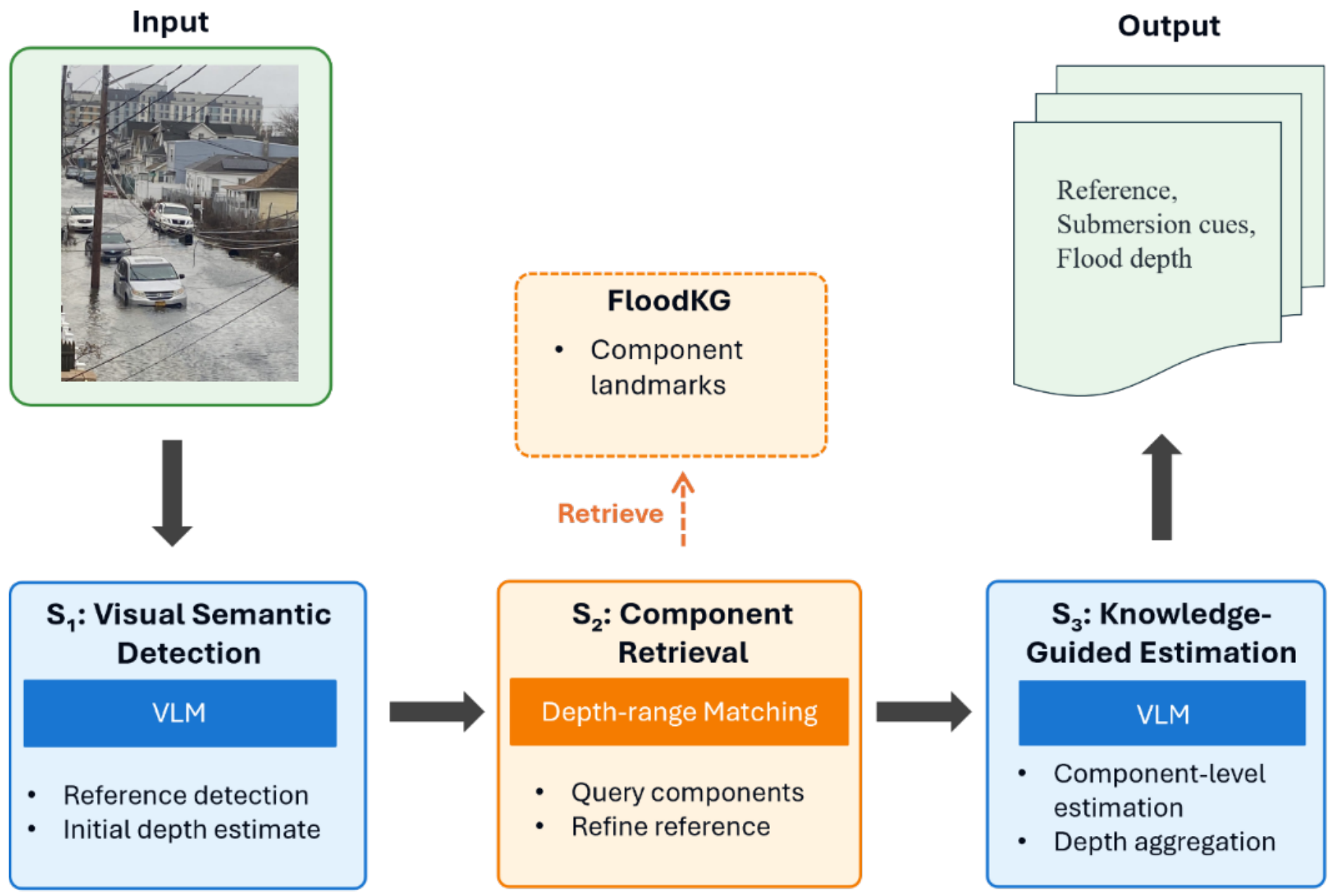

**Figure 1. Overview of the FloodVision framework for urban flood depth estimation.**

**Domain Knowledge Base.** FloodKG provides geometric priors for objects frequently observed in flooded urban environments. The knowledge base includes eight object types grouped into three categories: vehicles, persons, and urban infrastructure. Each object type is associated with a canonical height range and a set of flood-relevant vertical components. A key design feature of FloodKG is its component-level representation. For each object type, FloodKG encodes vertical landmarks that remain visually interpretable under partial submersion, such as the bumper bottom or rocker panel for vehicles, and the ankle or knee for pedestrians. Compared to treating vehicles or persons as whole objects, these components provide finer-grained and more robust reference cues, particularly when full object visibility is unavailable. As summarized in Table 1, FloodKG contains a total of 63 component landmarks and is stored in a hierarchical JSON structure for efficient lookup during inference.

**Table 1. Structure of the FloodKG knowledge base, including object categories, and component-level landmarks.**

| Category | Object Type | Components |
|---|---|---|
| Vehicle | Sedan, SUV | Wheel_Bottom, Tread_WheelBottom_to_CenterCap, Rocker_Panel, Center_Cap, Tread_CenterCap_to_WheelArchTop, Wheel_Arch_Top, Bumper_Bottom, Headlight_Center, Door_Handle, Window_Bottom, Window_Top, Roof_Line |
| Person | Adult, Child | Foot_Sole, Ankle_Malleolus, Mid_Calf, Knee_Patella, Mid_Thigh, Hip_GreaterTroch, Waistline, Navel, Chest_NippleLine, Shoulder_Acromion, Elbow_Flexed, Wrist, Neck_Base, Chin, Eye_Line, Top_Head |

| Category | Object Type | Components |
|---|---|---|
| | Curb | Curb_Top |
| Infrastructure | Fire_Hydrant | Base_Flange, Barrel_Lower, Side_Nozzle, Cap_Flange, Top_Cap |
| | Traffic_Cone | Cone_Top |

**Visual Semantic Detection.** Given an input image $I$, Stage 1 aims to identify semantic anchors with recognizable geometry that can serve as vertical references. The VLM is used to detect such objects and output, for each detected object $i$, a natural-language description $o_i$, an estimated real-world height $h_i^{(1)}$, a submersion ratio $\rho_i^{(1)} \in [0,1]$, and an implied flood depth

$$d_i^{(1)} = h_i^{(1)} \cdot \rho_i^{(1)}.$$

This stage provides an object-level estimate of flood depth based on the VLM's internal world knowledge. However, these initial estimates lack explicit geometric grounding, motivating the knowledge-guided refinement in subsequent stages.

**Knowledge-Guided Component Retrieval.** The object descriptions produced in Stage 1 are free-form and may not correspond directly to canonical entries in FloodKG. Stage 2 therefore performs a normalization step that aligns each description $o_i$ to a canonical object type in the knowledge base. If no suitable match is found, the object proceeds without knowledge constraints. For matched objects, FloodKG is queried to retrieve flood-relevant components whose vertical positions are plausibly near the preliminary depth estimate. The selected components are summarized into a compact textual context $K$ for use in Stage 3.

**Component-Level Depth Estimation.** In Stage 3, the VLM is invoked with the original image $I$ , conditioned on the knowledge context $K$ derived from FloodKG. The model integrates visual evidence with the provided geometric priors and focuses on the selected components when estimating the water depth. For each component $c_j$, the VLM predicts how much of the component's vertical extent is submerged, and this submersion ratio is combined with the component's canonical height range to obtain a component-level depth estimate $d_j^{(2)}$. The final flood depth is obtained by aggregating multiple component-level estimates across the scene, summarized by their mean and range.

**Data.** In this study, we use flood-related imagery collected from the MyCoast New York platform (MyCoast New York 2025), a statewide community science initiative in which residents submit geotagged photographs documenting flooding events. From this repository, a dataset of 654 flood images was curated covering New York City and surrounding areas. These images are used as inputs to the proposed FloodVision framework, while the resident-reported flood depth estimates are treated as proxy ground truth for evaluation.

## EXPERIMENT AND RESULTS

**Experimental Setup.** FloodVision is implemented in Python and interacts with GPT-4o-mini through the OpenAI API (OpenAI 2024). Deterministic inference is enforced by setting the temperature to zero. FloodVision was evaluated using MAE and median absolute error to quantify deviations between predicted flood depths and proxy ground-truth measurements.

**Results and Analysis.** Table 2 summarizes the overall performance of the VLM-only baseline (Stage 1) and the knowledge-guided FloodVision framework (Stage 3). Across both metrics, FloodVision consistently outperforms the baseline. The MAE decreases from 15.62 cm to 8.75 cm, representing a 43.9% reduction, while the median absolute error drops from 14.35 cm to 7.75 cm (46.0% reduction). Because the error distribution is right-skewed (Figure 2), the median provides a more robust indicator of typical performance than the mean. The concurrent reductions in both metrics indicate that the performance gains extend beyond outlier suppression and apply to the majority of cases.

**Table 2. Overall performance of baseline (Stage 1) and knowledge-guided FloodVision (Stage 3).**

| Metric | Stage 1 | Stage 3 | Relative Improvement |
|---|---|---|---|
| MAE (cm) | 15.62 | 8.75 | 43.9% |
| Median Error (cm) | 14.35 | 7.75 | 46.0% |

The VLM-only baseline exhibits a broad, heavy-tailed error distribution, with a substantial fraction of images showing deviations exceeding 20 cm (Figure 2), which limits its reliability for operational applications. Incorporating FloodKG and component-level reasoning substantially narrows this distribution. Across the 654 images, 453 cases (69.3%) achieve lower absolute error in Stage 3 than in Stage 1, and the mean error for these improved cases decreases from 16.8 cm to 6.2 cm.

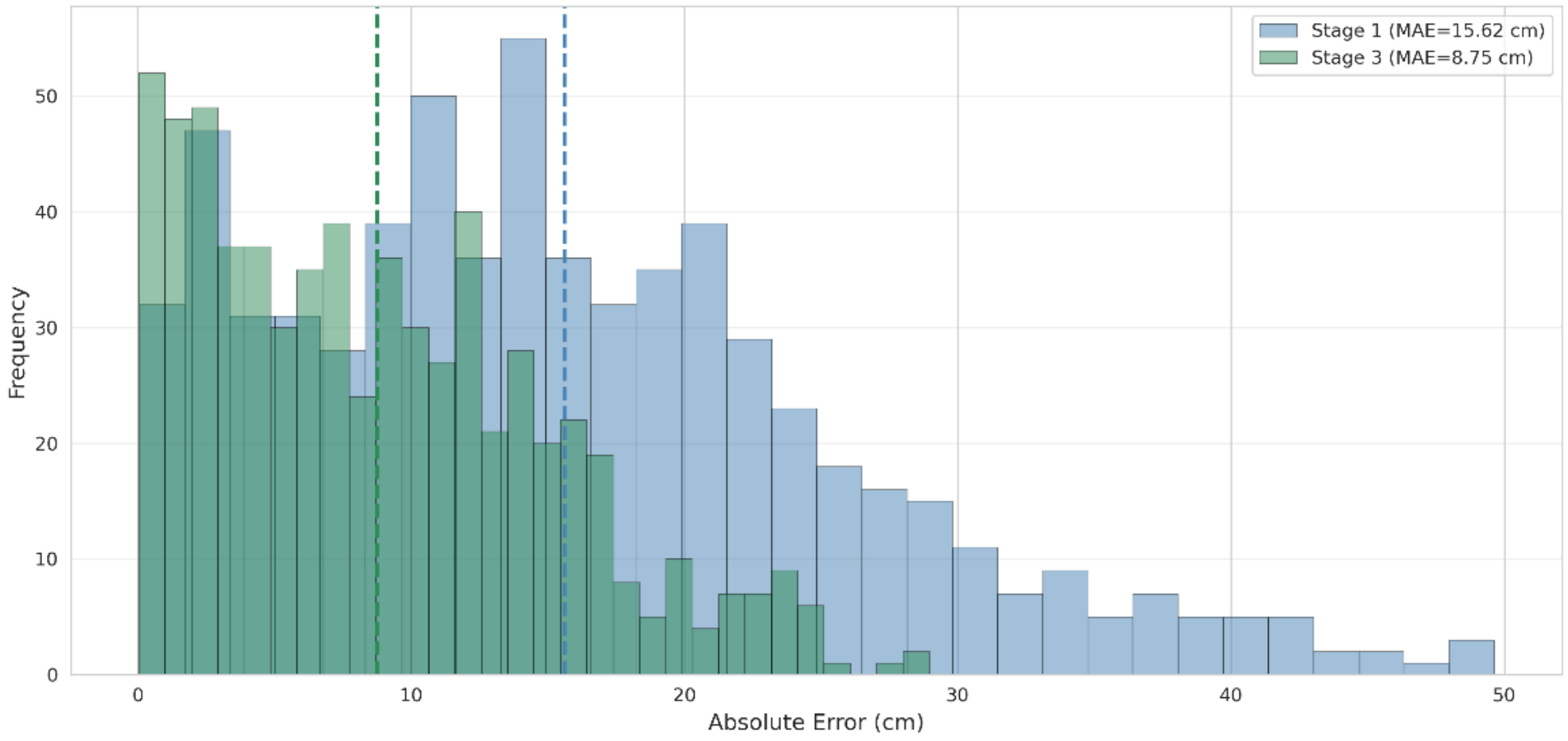

**Figure 2. Error distribution histogram for Stage 1 and Stage 3.**

## DISCUSSION

This study demonstrates that FloodVision improves image-based flood depth estimation by explicitly grounding VLM reasoning in component-level geometric priors. By integrating FloodKG into a staged inference process, FloodVision shifts from whole-object reasoning to fine-grained component-level reasoning. This design directs the model's attention toward geometrically meaningful landmarks that are directly related to water depth, thereby reducing ambiguity and improving estimation precision.

Compared with prior flood depth estimation approaches, FloodVision addresses key limitations in both supervised computer vision models and prompt-only VLM methods. Existing vision-based approaches have reported MAEs ranging from approximately 10 cm to over 30 cm (Alizadeh Kharazi and Behzadan 2021; Chaudhary et al. 2019), but they rely on large annotated datasets and are typically limited to predefined object categories during training, which restricts their generalization to unconstrained real-world scenes. More recent prompt-driven methods using GPT-4 achieve MAEs around 25 cm (Akinboyewa et al. 2024). In contrast, FloodVision injects geometric knowledge through FloodKG, enabling component-level grounding. On the real-world, crowdsourced MyCoast dataset, FloodVision reduces MAE from 15.62 cm to 8.75 cm and median error from 14.35 cm to 7.75 cm relative to a VLM-only baseline, indicating consistent gains across the majority of scenes (69.3%). From an application perspective, FloodVision can serve as a perception module for digital twins of urban flood systems, supporting road-level flood monitoring, inundation estimation across city-scale networks, and emergency response (Mohammadi and Taylor 2021).

Despite these strengths, several limitations remain. First, the current framework relies on the visibility of reference objects or components to infer water depth; in scenes with severe occlusion, estimation accuracy may degrade. Future work could incorporate additional visual signals, such as water surface texture, into FloodKG, and further leverage symbolic rule-based reasoning to enforce consistency and filter multiple candidate depth estimates (Xiong et al. 2025). In addition, the current FloodKG primarily encodes generic reference objects. Future extensions could enrich it with local urban characteristics to better capture regional variability in flood scenes. Moreover, synthetic urban flood datasets with predefined water depths may be explored for robustness evaluation, inspired by recent work on context-aware synthetic image generation (Lu et al. 2025). Finally, extending FloodVision to video streams would enable temporal consistency checks and short-term depth change analysis, providing richer support for flood monitoring and dynamic flood progression modeling.

## CONCLUSION

This paper presents FloodVision, a framework for urban flood depth estimation that integrates the semantic reasoning capabilities of foundation VLMs with geometric grounding from a curated knowledge base. By shifting from whole-object reasoning to component-level visual grounding, FloodVision enables more accurate flood depth estimation. Experimental results on 654 real-world, crowdsourced flood images demonstrate that FloodVision substantially improves estimation accuracy over a VLM-only baseline, reducing the MAE from 15.62 cm to 8.75 cm and the median error from 14.35 cm to 7.75 cm. FloodVision can be deployed with geotagged crowdsourced images or privacy-compliant CCTV streams for real-time flood impact assessment, supporting emergency routing, and road closure decisions. Future work will strengthen evaluation and assess robustness under controlled label noise and viewpoint variations.